\documentclass[runningheads]{llncs}
\usepackage{subfig}
\usepackage{tabularx}
\usepackage{array}
\usepackage{floatrow}
\usepackage{wrapfig}
\usepackage{verbatim}
\usepackage[font=small]{caption}
\usepackage{url}
\usepackage{hyperref}
\usepackage{cite}
\usepackage{longtable}
\usepackage{multirow}
\usepackage{pifont}

\newcolumntype{x}[1]{>{\raggedleft\arraybackslash\hspace{0pt}}p{#1}}
\newcolumntype{z}[1]{>{\centering\arraybackslash\hspace{0pt}}p{#1}}

\newcommand{\cmark}{\textcolor{black}{\ding{51}}}%
\newcommand{\xmark}{\textcolor{red}{\ding{55}}}%
\makeatletter
\newcommand*{\rom}[1]{\expandafter\@slowromancap\romannumeral #1@}

\usepackage{paralist}
\usepackage{graphicx}
\usepackage{amsmath,amssymb} % define this before the line numbering.
\usepackage[dvipsnames]{xcolor}
\usepackage{multirow}
\usepackage[width=122mm,left=12mm,paperwidth=146mm,height=193mm,top=12mm,paperheight=217mm]{geometry}
\usepackage[]{algorithm2e}
\usepackage[noend]{algpseudocode}
\makeatletter
\def\BState{\State\hskip-\ALG@thistlm}
\makeatother
\usepackage{booktabs}
\usepackage{sidecap}

\begin{document}
% \linenumbers
\pagestyle{headings}
\mainmatter

\title{It's Moving! A Probabilistic Model for Causal Motion Segmentation in Moving Camera Videos}
%\titlerunning{ECCV-16 submission ID \ECCV16SubNumber}

%\authorrunning{ECCV-16 submission ID \ECCV16SubNumber}

\author{Pia Bideau, Erik Learned-Miller}
\institute{University of Massachusetts, Amherst}

\maketitle

\begin{abstract}

The human ability to detect and segment moving objects works in the
presence of multiple objects, complex background geometry, motion of
the observer, and even camouflage. In addition to all of this, the
ability to detect motion is nearly instantaneous.  While there has
been much recent progress in motion segmentation, it still appears we
are far from human capabilities. In this work, we derive from first
principles a new likelihood function for assessing the probability of
an optical flow vector given the 3D motion direction of an object. This
likelihood uses a novel combination of the angle and magnitude of the
optical flow to maximize the information about the true motions of
objects. Using this new likelihood and several innovations in
initialization, we develop a motion segmentation algorithm that beats
current state-of-the-art methods by a large margin. We compare to 
five state-of-the-art methods on two established
benchmarks, and a third new data set of camouflaged animals, which
we introduce to push motion segmentation to the next level.

\end{abstract}

\vspace{-.2in}
{\em ``Motion is a powerful cue for image and scene segmentation in
  the human visual system. This is evidenced by the ease with which we
  see otherwise perfectly camouflaged creatures as soon as they
  move.''} --Philip Torr~\cite{torr1998geometric}

\begin{SCfigure}
   \includegraphics[width=0.5\linewidth]{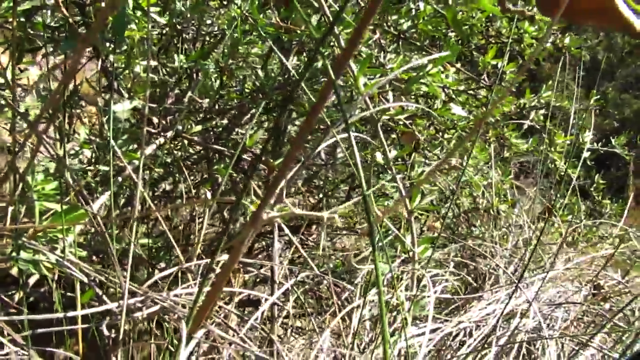}
   \caption{\textbf{Where is the camouflaged insect?} Before looking
     at Figure~\ref{fig:phasmid_gt}, which shows the ground truth
     localization of this insect, try identifying the insect. While it
     is virtually impossible to see without motion, it immediately
     ``pops out'' to human observers as it moves in the video (see supplementary
     material).}
\label{fig:phasmid_plain}
\end{SCfigure}

\vspace{-.2in}

\section{Introduction}
How can we match the ease and speed with which humans and other
animals detect motion? This remarkable capability works in the
presence of complex background geometry, camouflage, and motion of the
observer. Figure~\ref{fig:phasmid_plain} is a frame from a video of a
``walking stick'' insect. Despite the motion of the camera, the rarity 
of the object, and the high complexity of the
background geometry, the insect is immediately visible as soon as it
starts moving.

To develop such a {\em motion segmentation} system, we re-examined
classical methods based upon perspective projection, and developed a
new probabilistic model which accurately captures the information
about 3D motion in each optical flow vector. In particular, we derive
a new {\em conditional flow angle likelihood},
$p(\mathbf{t}_\theta|M,\mathbf{t}_r)$, the probability of observing a
particular flow angle $\mathbf{t}_\theta$ given the flow magnitude
$\mathbf{t}_r$ and the 3D motion direction $M$ of an object (for
brevity, we will refer to it as the {\em angle likelihood}). This
angle likelihood is derived from the fundamental perspective
projection image formation model, and a model of optical flow as a
noisy observation of (2D) scene motion.

This new angle likelihood helps us to address a fundamental difficulty
of motion segmentation: the ambiguity of 3D motion given a set of flow
vectors.\footnote{Ogale et al.~\cite{ogale2005motion} referred to the
  set of 3D motions compatible with a particular set of optical flow
  vectors as the {\em motion valley}, due to its typical appearance as
  a long narrow region on the surface of a sphere defining possible
  motion directions in 3D.} While we cannot eliminate this problem
completely, the angle likelihood allows us to weigh the evidence for
each image motion properly based on the optical flow. In particular,
when the underlying image motion is very small, moderate errors in the
optical flow can completely change the apparent motion direction
(i.e., the angle of the optical flow vector). When the underlying
image motion is large, typical errors in the optical flow will not
have a large effect on apparent motion direction. This leads to the
critical observation that {\em small optical flow vectors are less
  informative about motion than large ones.} Our derivation of the
angle likelihood (Section~\ref{sec:blind}) quantifies this
notion and makes it precise in the context of a Bayesian model of
motion segmentation.

\begin{SCfigure}
   \includegraphics[width=0.5\linewidth]{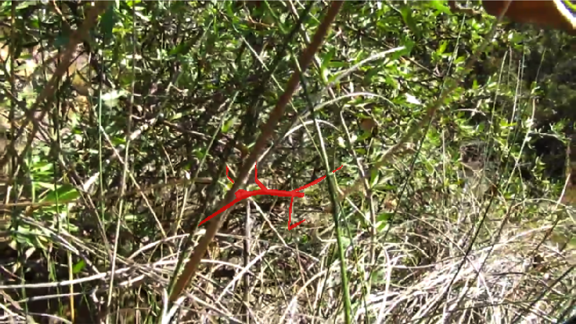}
   \caption{\textbf{Answer:} the insect from Figure~1 in shown in red. The
     insect is trivial to see in the original video, though extremely
     difficult to identify in a still image. In addition to superior results
on standard databases, our method is also one of the few that can detect 
objects is such complex scenes.}
\label{fig:phasmid_gt}
\end{SCfigure}

We evaluate our method on three diverse data sets, achieving 
state-of-the-art performance on all three. The first is the widely
used Berkeley Motion Segmentation (BMS-26) database~\cite{Brox10,Tron07},
featuring videos of cars, pedestrians, and other common scenes. The
second is the Complex Background Dataset~\cite{Narayana13}, designed
to test algorithms' abilities to handle scenes with highly variable
depth. Third, we introduce a new and even more challenging
benchmark for motion segmentation algorithms: the {\em Camouflaged
  Animal Data Set}. The nine (moving camera) videos in this benchmark
exhibit camouflaged animals that are difficult to see in a single
frame, but can be detected based upon their motion across frames.
Our dominance on these three substantially different data sets exhibits 
the high quality of our method. 

\section{Related Work}
A large number of motion segmentation approaches have been proposed,
including~\cite{grundmann2010efficient,lezama11,kumar2008learning,Irani94,Ren03,Sheikh09,Elqursh12,Ochs12,Kwak11,Brox10,rahmati2014motion,jain2014supervoxel,Manju,Zamalieva,Ferrari,keuper2015motion,taylor2015causal,fragkiadaki2012video,sawhney1999independent,dey2012detection,Namdev2012}.
The prior literature is too large to review here, so we focus on
recent methods.

Many methods for motion segmentation work by tracking points or
regions through multiple frames to form motion trajectories, and
grouping these trajectories into coherent moving
objects~\cite{keuper2015motion,Brox10,Ferrari,fragkiadaki2012video,elhamifar2009sparse}. Elhamifar
and Vidal~\cite{elhamifar2009sparse} track points through multiple
images and show that rigid objects are represented by low-dimensional
subspaces in the space of tracks. They use sparse subspace clustering
to identify separate objects. Brox and Malik~\cite{Brox10} define a
pairwise metric on multi-frame trajectories so that they may be
clustered to perform motion segmentation.  Fragkiadaki et
al.~\cite{fragkiadaki2012video} detect discontinuities of the {\em
  embedding density} between spatially neighboring trajectories.
These discontinuities are used to infer object boundaries and perform
segmentation.  Papazoglou and Ferrari~\cite{Ferrari} develop a method
that looks both forward and backward in time, using flow angle and
flow magnitude discontinuities, appearance modeling, and superpixel
mapping across images to connect independently moving objects across
frames. Keuper et al.~\cite{keuper2015motion} also track points across
multiple frames and use minimum cost multicuts to group the
trajectories.

Note that these trajectory-based methods are {\em non-causal}. 
To segment earlier frames, they must wait for
trajectories which are computed over future frames. Our method,
however, is causal, relying only on the flow between two
frames and information passed forward from previous frames. Despite
this, we outperform trajectory-based methods by a large margin (see
Experiments).

Another set of methods analyze optical flow
between a pair of frames, grouping pixels into regions whose flow is
consistent with various motion models.
Torr~\cite{torr1998geometric} develops a sophisticated probabilistic
model of optical flow, building a mixture model
that explains an arbitrary number of rigid components within the
scene. Interestingly, he assigns different types of motion models to
each object based on model fitting criteria. His approach is
fundamentally based on projective geometry rather based directly on
perspective projection equations, as in our approach. Horn has
identified drawbacks of using projective geometry in such
estimation problems and has argued that methods based directly on
perspective projection are less prone to overfitting in the presence
of noise~\cite{horn1999projective}.  Zamalieva et al.~\cite{Zamalieva}
present a combination of methods that rely on homographies and
fundamental matrix estimation. The two methods have complimentary
strengths, and the authors attempt to select among the best
dynamically. An advantage of our method is that we do not depend upon
the geometry of the scene to be well-approximated by a group of
homographies, which enables us to address videos with very complex
background geometries.  Narayana et al.~\cite{Manju} remark that for
translational only motions, the {\em angle field} of the optical flow
will consist of one of a set of canonical angle fields, one for each
possible motion direction, regardless of the focal length. They use
these canoncial angle fields as a basis with which to segment a motion
image. However, they do not handle camera rotation, which is a
significant limitation.

Another set of methods using occlusion events in video to reason about
depth ordering and independent object motion~\cite{ogale2005motion,taylor2015causal}. Ogale et al.~\cite{ogale2005motion} use occlusion cues to further
disambiguate non-separable solutions to the motion segmentation problem.
Taylor et al.~\cite{taylor2015causal} introduce a causal framework for
integrating occlusion cues by exploiting temporary consistency priors to
partition videos into depth layers.

Estimation of a camera's translation and rotation from the observed
optical flow is closely related to motion
segmentation~\cite{Jepson94,Prazdny83,Tomasi93,Prazdny80,Tian96,Yamaguchi13,hartley1997defense}. For
lack of space, we discuss these works in Supp. Mat.

\section{Methods}
\label{sec:blind}

The {\em motion field} of a scene is created by the movement of the
camera relative to a stationary background and the additional motion
of independently moving objects. We use the optical flow,
or estimated motion field, to segment each video image into background
and other independently moving objects.

When the camera is only translating (and not rotating) relative to the
background, there are strong constraints on the background's optical
flow--the {\em direction} or {\em angle} $\vec{t_{\theta}}$ of the motion at each pixel is determined by the camera translation $(U,V,W)$, the image location
of the pixel $(x, y)$, and the camera's focal length $f$, and has no dependence on
scene depth~\cite{RobotVision}. 
\begin{eqnarray}
\vec{t_\theta} = \arctan(W\cdot y - V\cdot f, W\cdot x - U\cdot f)
\end{eqnarray}

%As a result, given several hypotheses
%about how a point in the scene is moving, an optical flow vector is
%frequently enough information to guess the correct motion hypothesis.
{\bf Simultaneous camera rotation and translation}, however, couple
the scene depth and the optical flow, making it much harder to assign
pixels to the right motion model.

To address this, we wish to {\em subtract off the estimated rotational
  component $\hat{O}_R$ of optical flow} from the original flow $O$ to
produce a translation component estimate~$\hat{O}_T$.  The subsequent
assignment of flow vectors to motion models is thus greatly
simplified. However estimating camera rotation in the presence of multiple
motions is challenging. We organize the Methods section as follows.

In Section~\ref{sec:model}, we describe how all frames after the first frame
are segmented, using the segmentation from the previous frame and our
novel angle likelihood.  After reviewing Bruss and Horn's motion estimation
technique~\cite{bruss1983passive} in Section~\ref{sec:cam_est}, Section~\ref{sec:init} 
describes how our method is initialized in the first frame, including
a novel process for estimating camera motion in the presence of multiple
motions.

\subsection{A probabilistic model for motion segmentation}
\label{sec:model}
\begin{figure}
   \includegraphics[width=0.9\linewidth]{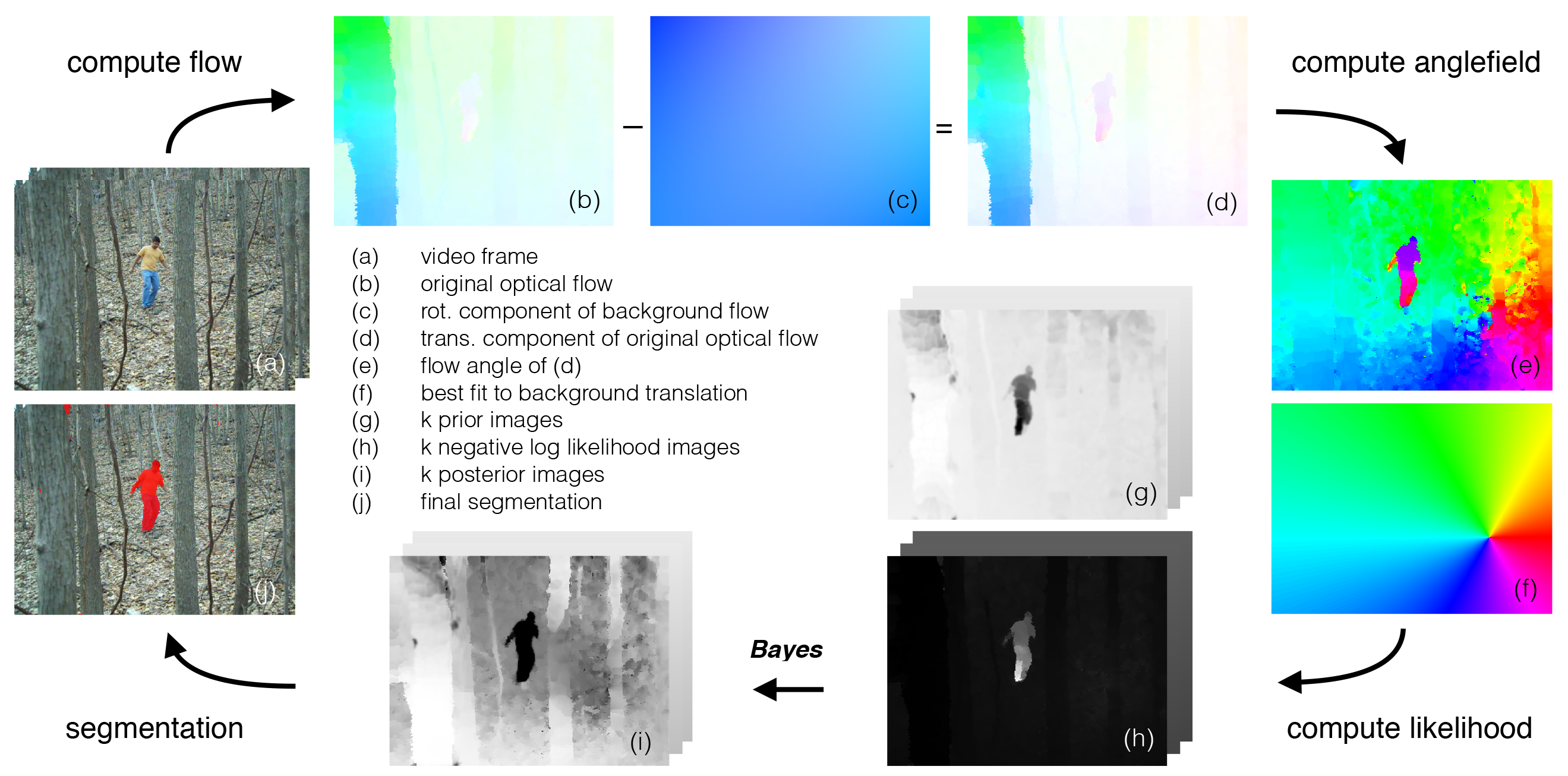}
   \caption{{\bf Our segmentation procedure.} Given the optical flow
     (b) the camera motion is estimated. Then, the rotational flow
     $\hat{O}_R$ (c) is subtracted from the optical flow $O$ to
     produce a translational flow $\hat{O}_T$. The angle field of
     $\hat{O}_T$ is shown in (e). The best fitting translation
     parameters to the background of $\hat{O}_T$ yield an angle field
     (f), which clearly shows the forward motion of the camera
     (rainbow focus of expansion pattern) not visible in the original
     angle field. The motion component priors (g) and negative log
     likelihoods (h) yield the posteriors (i) and the final segmentation (j).}
\label{fig:forest}
\end{figure}

Given a (soft) motion segmentation of frame $T-1$ into $k$ moving
objects and the optical flow $O$ from frames $T$ and $T+1$, segmenting
frame $T$ requires several ingredients: {\bf a)} the {\em prior}
probabilities $p(M_j)$ for each pixel that it is assigned to each
particular motion model $M_j$, {\bf b)} an estimate of the current
3D motion directions $M_j$, $1\leq j\leq k$, for each of the $k$ objects
from the previous frame, {\bf c)} for each pixel position, an {\em
  angle likelihood} $p(\mathbf{t}_\theta|M_j,\mathbf{t}_r)$ that gives the likelihood of each flow angle
$\mathbf{t}_\theta$ under each motion model $M_j$ conditioned on the
flow magnitude $\mathbf{t}_r$, and {\bf d)} the prior probability
$p(M_{k+1})$ and angle likelihoods
$p(\vec{t}_\theta|M_{k+1},\vec{t}_r)$ of a new motion $M_{k+1}$. Given
these priors and likelihoods, we simply use Bayes' rule to obtain
posterior probabilities of each motion at each pixel location. We have
\begin{eqnarray}
\label{eq:Bayes}
p(M_j|\vec{t}_\theta,\vec{t}_r)&\propto & p(\vec{t}_\theta|M_j,\vec{t}_r)\cdot p(M_j|\vec{t}_r)\\
&=& p(\vec{t}_\theta|M_j,\vec{t}_r)\cdot p(M_j).
\end{eqnarray}
The second expression follows since the prior $M_j$ does not depend on
$\vec{t}_r$.  We directly use this posterior for segmentation.  We now
describe how the above quantities are computed.

{\bf Propagating the posterior for a new prior.}  We start from the
optical flow of Sun et al.~\cite{sun2010secrets}
(Figure~\ref{fig:forest}b). We then create a {\em prior} at each pixel
for each motion model in the new frame (Figure~\ref{fig:forest}g) by
propagating the posterior from the previous frame
(Figure~\ref{fig:forest}i) in three steps.
\begin{itemize}
\item Use the previous frame's flow to map posteriors
  from frame $T-1$ (Figure~\ref{fig:forest}i) to new positions in frame $T$.
\item Smooth the mapped posterior in the new frame by convolving with
  a spatial Gaussian, as done in
  ~\cite{Manju,narayana2014background}. This implements the idea that
  object locations in future frames are likely to be close to their
  locations in previous frames.
\item Renormalize the smoothed posterior from the previous frame to
  form a proper probability distribution at each pixel location, which
  acts as the prior on the $k$ motion components for the new frame
  (Figure~\ref{fig:forest}g). Finally, we set aside a probability of
  $1/(k+1)$ for the prior of a {\em new motion component}, while
  rescaling the priors for the pre-existing motions to sum to
  $k/(k+1)$.
\end{itemize}

{\bf Estimating and removing rotational flow.}  We use the prior for
the background component to weight pixels for estimating
the current frame's flow due to the camera.\footnote{Note that we do
  not explicitly estimate the camera motion, since we know neither the
  speed nor focal length of the camera, only the direction of motion.}
We estimate the camera translation parameters $(U,V,W)$ and rotation
parameters $(A,B,C)$ using a modified version of the Bruss and Horn
algorithm~\cite{bruss1983passive}~(Section~\ref{sec:cam_est}).  As
described above, we then render the flow {\bf angle} {\em independent
  of the unknown scene depth} by subtracting the estimated rotational
component $\hat{O}_R$ (Figure~\ref{fig:forest}c) from the original flow
$O$ (Figure~\ref{fig:forest}b) to produce an estimate of the
translational flow $\hat{O}_T$ (Fig.~\ref{fig:forest}d):
\begin{equation}
\hat{O}_T=O-\hat{O}_R(\hat{A},\hat{B},\hat{C}).
\end{equation} We denote by $\vec{t}$ the estimated translation at
a particular pixel.
Next, for each addtitional motion component, we estimate 3D translation direction
parameters using the segment priors to select pixels, weighted
according to the prior. Let the collection of translation parameters
for motion component $j$ be denoted $M_j$.

{\bf The flow angle likelihood.}  Once we have obtained $\hat{O}_T$ by
removing the rotational flow, we use each flow vector $\vec{t}$ to
decide which motion component it belongs to. Most of the
information about the 3D motion direction is contained in the {\em
  flow angle}, not the flow magnitude. This is because for a given
translational 3D motion direction (relative to the camera), the flow
angle is {\em completely determined} by that motion and the location
in the image, whereas the flow magnitude is a function of the object's
depth, which is unknown. However, as discussed above, the {\em amount
  of information} in the flow angle depends upon the flow
magnitude--flow vectors with greater magnitude are much more reliable
indicators of true motion direction. This is why it is critical to
formulate the angle likelihood conditioned on the flow magnitude.

Other authors have used flow angles in motion
segmentation.  For example, Papazoglou and Ferrari~\cite{Ferrari} use
both a gradient of the optical flow and a separate function of the
flow angle to define motion boundaries.  Narayana et al.~\cite{Manju}
use {\em only} the optical flow angle to
evaluate motions. But our derivation gives a principled, novel, and
highly effective method of using the flow angle and magnitude together
to mine accurate information from the optical flow.  In particular, we
show that while (under certain mild assumptions) the translational
magnitudes alone have no information about which motion is most
likely, the magnitudes play an important role in specifying the {\em
  informativeness} of the flow angles.  In our experiments section, we
demonstrate that failing to condition on flow magnitudes in this way
results in greatly reduced performance over our derived model.

We now derive the most important element of our method, a high quality
{\em conditional flow angle likelihood}
$p(\vec{t}_\theta|M_j,\vec{t}_r)$, the probability of observing a flow
direction $\vec{t}_\theta$ given that a pixel was part of an object
undergoing motion $M_j$, and that the flow magnitude was $\vec{t}_r$.
We make the following modeling assumptions:
\vspace{10pt}
\begin{compactenum}
\item We assume the observed translational flow 
$\vec{t}=(\vec{t}_r,\vec{t}_\theta)$ at a pixel is a noisy
observation of the unobserved translational motion field
$\vec{t}^*=(\vec{t}_r^*,\vec{t}_\theta^*)$:
\begin{equation}
\label{eq:noise}
\vec{t}=\vec{t}^* + \eta,
\end{equation}
 where $\eta$ is independent 2D Gaussian noise with
0-mean and circular but unknown covariance $s\cdot I$.  
\item We assume the translational motion field magnitude $\vec{t}_r^*$
  is statistically independent of the translation motion field angle
  $\vec{t}_\theta^*$. It follows that $\vec{t}_r=\vec{t}_r^*+\eta$ is
  also independent of $\vec{t}_\theta^*$, and hence
  $p(\vec{t}_r|\vec{t}_\theta^*)=p(\vec{t}_r)$.
\item A motion model $M_j$ gives the direction of 3D motion, but not
  its magnitude. For a given position in the image, each 3D 
motion direction yields a (2D) motion field direction $\vec{t}_\theta^*$
  in the image.  We assume that $\vec{t}_\theta^*$ contains all of the
  information about the motion model useful for predicting the optical
  flow, or $p(\vec{t}|M_j) = p(\vec{t}|\vec{t}_\theta^*)$.
\end{compactenum}
\vspace{10pt}
With these assumptions, we have
\begin{eqnarray}
p(\vec{t}|M_j)
&\overset{(3)}{=}&p(\vec{t}|\vec{t}_\theta^*)\\
&\overset{(1)}{=}&p(\vec{t}_r,\vec{t}_\theta|\vec{t}_\theta^*)\\
&=&p(\vec{t}_\theta|\vec{t}_r,\vec{t}_\theta^*)\cdot p(\vec{t}_r|\vec{t}_\theta^*)\\
&\overset{(2)}{=}&p(\vec{t}_\theta|\vec{t}_r,\vec{t}_\theta^*)\cdot p(\vec{t}_r)\\
\label{ref:propto}&\propto & p(\vec{t}_\theta|\vec{t}_r,\vec{t}_\theta^*)\\
&\overset{(3)}{=}&p(\vec{t}_\theta|M_j,\vec{t}_r),
\end{eqnarray}
where the numbers over each equality give the assumption that is invoked.
Equation~(\ref{ref:propto}) follows since $p(\vec{t}_r)$ is constant across all motion models.

\begin{figure}[t]
   \includegraphics[width=0.98\columnwidth]{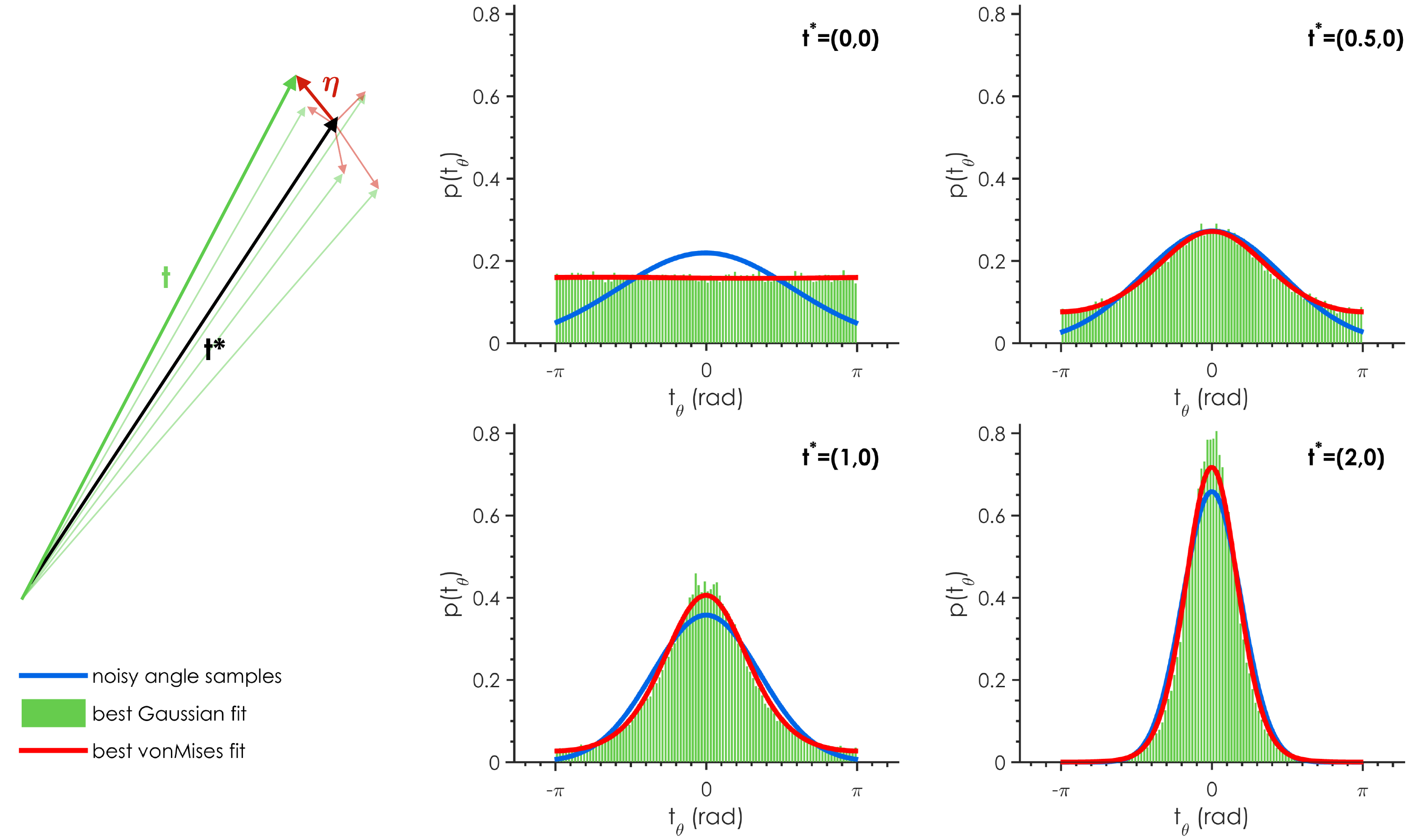}
   \caption{{\bf The von Mises distribution.} When a motion field
     vector $\vec{t}^*$ is perturbed by added Gaussian noise $\eta$ (figure top left), the resulting distribution over optical flow angles
     $\vec{t}_\theta$ is well-modeled by a {\em von Mises}
     distribution. The figure shows how small motion field vectors
     result in a broad distribution of angles after noise is
     added, while larger magnitude motion field vectors result
     in a narrower distribution of angles. The red
     curve shows the best von Mises fit to these sample distributions
     and the blue curve shows the lower quality of the best Gaussian
     fit.}
\label{fig:vonMises}
\end{figure}

We model $p(\vec{t}_\theta|\vec{t}_r,\vec{t}_\theta^*)$ using a {\em
  von Mises} distribution with parameters $\mu$, the preferred direction,
and concentration parameter $\kappa$. We set $\mu=\vec{t}_\theta^*$, since the
most likely observed angle $\vec{t}_\theta$ is the ground truth angle
$\vec{t}_\theta^*$. To set $\kappa$, we observe that when the {\em
  ground truth} flow magnitude $\vec{t}_r^*$ is small, the
distribution of observed angles $\vec{t}_\theta$ will be near uniform
(see Figure~\ref{fig:vonMises}, $\vec{t}^*=(0,0)$), whereas when
  $\vec{t}_r^*$ is large, the observed angle $\vec{t}_\theta$ is
  likely to be close to the ground truth flow angle $\vec{t}_\theta^*$
  (Figure~\ref{fig:vonMises}, $\vec{t}^*=(2,0)$). We can achieve this basic relationship by
    setting $\kappa=a (\vec{t}_r^*)^b$, where $a$ and $b$ are
    parameters that give added flexibility to the model. Since we
    don't have direct access to $\vec{t}_r^*$, we use $\vec{t}_r$ as a
    surrogate, yielding
\begin{equation}
%p(\vec{t}_i|M^i_j,\vec{t}_r)\propto \mbox{vonMises}(\theta_i; \mu=\theta^*_i, \kappa=a r_i^b).
p(\vec{t}|M_j)\propto \mbox{vonMises}(\vec{t}_\theta; \mu=\vec{t}_\theta^*, \kappa=a {\vec{t}_r}^b).
\end{equation}
Note that this likelihood treats zero-length translation vectors
as uninformative--it assigns them the same likelihood under all
motions. This makes sense, since the direction of a zero-length optical
flow vector is essentially random. Similarly, the longer the optical
flow vector, the more reliable and informative it becomes.

{\bf Likelihood of a new motion.}  Lastly, with no prior
information about new motions, we set
$p(\vec{t}_\theta|M_{k+1})=\frac{1}{2\pi}$, a uniform distribution.

Once we have priors and likelihoods, we compute the posteriors
(Equation~\ref{eq:Bayes}) and label each pixel as
\begin{equation}
L=\underset{j}{\arg\max} \;p(M_j|\vec{t}_\theta,\vec{t}_r).
\end{equation} 

\subsection{Bruss and Horn's motion estimation.}
\label{sec:cam_est}
To estimate the {\em direction} of motion (but not the speed) of the
camera relative to the background, we use the method of Bruss and
Horn~\cite{bruss1983passive} and apply it to pixels selected by the
background prior.  The optical flow vector $\vec{v}_i$ at
pixel $i$ can be decomposed as $\vec{v}_i=\vec{p}_i+\vec{e}_i$, where
$\vec{p}_i$ is the component of $\vec{v}_i$ in the direction predicted
by the motion model and $\vec{e}_i$ is the component orthogonal to
$\vec{p}_i$.  The authors find the motion $M$ that minimizes the sum
of these ``error'' components $\vec{e}_i$. The optimization for 
{\bf translation-only} is
\begin{equation}
\underset{U,V,W}{\arg \min} \sum_i \|\vec{e}_i(\vec{v}_i,U,V,W)\|,
\end{equation}
where $U$, $V$, and $W$ are the three translational motion components.
Bruss and Horn give a closed form solution to this problem for the
translation-only case. 

{\bf Recovering camera rotation.}  Bruss and Horn also outline how to
solve for rotation, but give limited details.  We implement our own
estimation of rotations $(A,B,C)$ and translation as a nested
optimization:
\begin{equation}
\label{eq:opt}
\hat{M}=\underset{A,B,C,U,V,W}{\arg\min} \left[\underset{U,V,W}{\min}\; 
\sum_i\|\vec{e}_i\left(\vec{v}_i,A,B,C,U,V,W\right)\| \right].
\end{equation}

Given the portion $\hat{O}_R$ of the observed flow $O$ due to
rotation, one can subtract off the rotation since it
does not depend on scene geometry: $\hat{O}_T=O-\hat{O}_R$.

Subtracting the rotation $(A,B,C)$ from the observed flow
reduces the optimization to the translation only case.
We solve the optimization over the rotation parameters
$A,B,C$ by using Matlab's standard gradient descent optimization,
while calling the Bruss and Horn closed form solution for the
translation variables given the rotational variables as part of the
internal function evaluation. Local minima are a concern, but since we
are estimating camera motion between two video frames, the rotation is
almost always small and close to the optimization's starting
point. 

\subsection{Initialization: Segmenting the first frame}
\label{sec:init}
The goals of the initialization are a) estimating background
translation and rotation parameters, b) finding pixels whose flow is
consistent with this motion model, and c) assigning inconsistent
groups of contiguous pixels to additional motion models.  Bruss and
Horn's method was not developed to handle scenes with multiple
different motions, and so large or fast-moving foreground objects can result
in poor motion estimates (Figure~\ref{fig:RANSAC_comp}).

\begin{figure}
   \includegraphics[width=0.28\columnwidth]{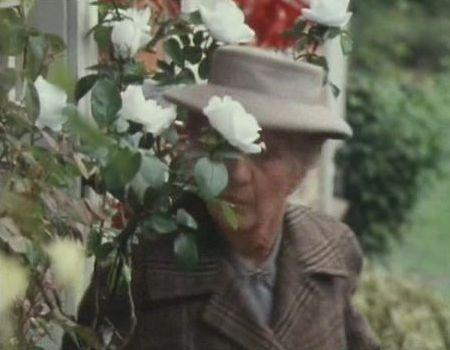}
   \includegraphics[width=0.28\columnwidth]{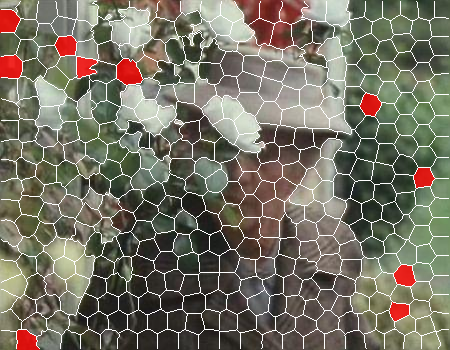}
   \includegraphics[width=0.28\columnwidth]{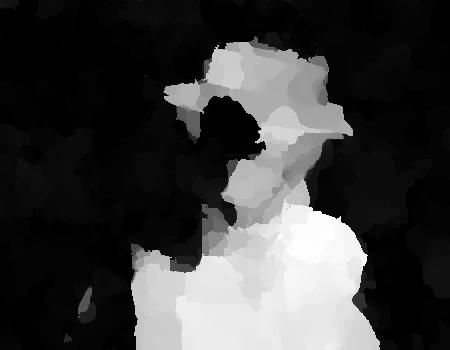}
   \caption{{\bf RANSAC procedure.} The result of our RANSAC procedure
     to find background image patches. Notice that none of the patches
     are on the person moving in the foreground. Also notice that we
     force the algorithm to pick patches in three of the four image
     corners (a ``corner'' is 4\% of the image). The right figure
     shows the negative log likelihood of background.}
\label{fig:RANSAC}
\end{figure}

{\bf Constrained RANSAC}. To address this problem we use a modified
version of RANSAC~\cite{RANSAC} to robustly estimate background motion
(Figure~\ref{fig:RANSAC}). We use 10 random SLIC
superpixels~\cite{SLIC}\footnote{We use the
  http://www.vlfeat.org/api/slic.html code with regionSize=20 and
  regularizer=0.5.} to estimate camera motion
(Section~\ref{sec:cam_est}). We modify the standard RANSAC procedure
to force the algorithm to choose three of the 10 patches from the
image corners, because image corners are prone to errors due to a misestimated camera rotation.
Since the Bruss and Horn error function (Equation~\ref{eq:opt}) does not
penalize motions in a direction opposite of the predicted motion, we
modify it to penalize these motions appropriately (details in
Supp.~Mat.).  5000 RANSAC trials are run, and the camera motion
$\hat{M}$ resulting in the fewest outlier pixels according to the {\em modified Bruss-Horn} (MBH) error is retained, using a threshold of 0.1.

%\vspace{5pt}
\begin{algorithm}[H]
 \KwIn{video with $n$ frames}
 \KwOut{binary motion segmentation}
 %$n \gets $ number of RANSAC iterations\;
 %$m \gets $ number of pixels per frame\;
 %$frames \gets $ number video frames\;
 \For{$T \gets$ 1 \textbf{to} $n-1$} {
 compute opticalflow from frame $T$ to frame $T+1$\;
 \eIf{first frame}{
	 \ForEach{RANSAC iteration} {
	   find best motion model for 10 random patched (3 in corners)\;
	   retain best motion model for the static backgound $M_b$\; 
	 }
	 $p(M_j), p(M_b) \gets$ segment MBH error image into $k$ components using Otsu's method\;
    }{
	 $p(M_j), p(M_b) \gets$ propagate posterior $p(M_j|\vec{t}), p(M_b|\vec{t})$\;
	 compute motion model $M_b$ of static background\;
   }
 \ForEach{pixel in $\hat{O}_T$} {
 $p(\vec{t}_\theta|M_b,\vec{t}_r) \gets  \mbox{vonMises}(\vec{t}_\theta;\mu(\vec{t}_\theta^*),\kappa(\vec{t}_r))$\;
 %$p(M_b|\vec{t})\gets p(\vec{t}_\theta|M_b,\vec{t}_r)\cdot p(M_b)$\;
 }
 \For{$j \gets 1$ \textbf{to} k-1} {
    compute motion model $M_j$ of moving object $j$\;
    \ForEach{pixel in $\hat{O}_T$} {
        $p(\vec{t}_\theta|M_j,\vec{t}_r) \gets  \mbox{vonMises}(\vec{t}_\theta;\mu(\vec{t}_\theta^*),\kappa(\vec{t}_r))$\;
		%$p(M_j|\vec{t})\gets p(\vec{t}_\theta|M_j,\vec{t}_r)\cdot p(M_j)$\;  
      }
  }
  \ForEach{pixel in $\hat{O}_T$} {
   $p(M_{k+1}) \gets \frac{1}{k + 1}$\;
   $p(\vec{t}_\theta|M_{k+1},\vec{t}_r) \gets  \frac{1}{2\pi}$\;
   normalize $p(M_b)$ and $p(M_j)$ such that they sum up to $1-p(M_{k+1})$\;
   $p(M|\vec{t})\gets p(\vec{t}_\theta|M,\vec{t}_r)\cdot p(M)$\;
   %$p(M_{j}|\vec{t})\gets p(\vec{t}_\theta|M_{j},\vec{t}_r)\cdot p(M_{j})$\;
   %$p(M_{k+1}|\vec{t})\gets p(\vec{t}_\theta|M_{k+1},\vec{t}_r)\cdot p(M_{k+1})$\;
   }
  given the posteriors $p(M_b|\vec{t})$, $p(M_j|\vec{t})$ and $p(M_{k+1}|\vec{t})$ assign every pixel one of two labels: static background or moving objects\;
  }
 \caption{A causal motion segmentation algorithm}
\end{algorithm}
%\vspace{5pt}
{\bf Otsu's Method.} While using the RANSAC threshold on the MBH image
produces a good set of pixels with which to estimate the background
motion, the method often excludes some pixels that should be included
in the background motion component. We use Otsu's method~\cite{Otsu}
to separate the MBH image into a region of low error (background) and
high error:
\begin{itemize}
\item Use Otsu's threshold to divide the errors, minimizing the
  intraclass variance. Use this threshold to do a binary segmentation
  of the image.
\item Find the connected component $C$ with highest average error.
Remove these pixels ($I\gets I\setminus C$), and
assign them to an additional motion model.
\end{itemize}
These steps are repeated until Otsu's {\em effectiveness} parameter is
below 0.6.

\begin{comment}
\begin{algorithm}
\caption{Motion Segmentation}\label{MotionSegAlgorithm}
\begin{algorithmic}[1]
\Procedure{MotionSeg(video)}{}
\BState \emph{Process first frame (initialization)}:
\State Compute optical flow $O$ for first frame.
\State {\em Constrained RANSAC Minimization:}
\For{i=1 to 5000}
\State Find best motion model for 10 random patches (3 in corners).
\EndFor  
\State Retain best background motion model $\hat{M}$.
\State {\em Segment first frame:}
\State Calculate rotation component $\hat{O}_R$ from $\hat{M}$.
\State $\hat{O}_T\gets O-\hat{O}_R$  \Comment{Get translation component.}
\State Segment MBH error image with Otsu's method.
\BState \emph{Segment remaining frames}:
\For{each video frame}
  \State Compute optical flow
\State $\text{Prior}_{t} \gets \textit{Flow}(\text{Posterior}_{t-1}$)
\State Find motion $M_B$ of background component.
\State Estimate rotational field $\hat{O}_R$ from $M_B$.
\State $\hat{O}_T\gets O-\hat{O}_R$.
\For{each segment $j$}
\State {\em Compute motion model $M_j$:}
\State $M_j\gets BrussHorn(\text{Prior}_j,\hat{O}_T)$
\For{each vector $\vec{t}$ in $\hat{O}_T$}
\State $p(\vec{t}_\theta|M_j,\vec{t}_r)=\mbox{vonMises}(\vec{t}_\theta;\mu(\vec{t}_\theta^*),\kappa(\vec{t}_r))$
\State $p(M_j|\vec{t})\propto p(\vec{t}_\theta|M_j,\vec{t}_r)\cdot p(M_j)$
\Comment{Bayes.}
\EndFor
\EndFor
\EndFor
\EndProcedure
\end{algorithmic}
\end{algorithm}
\end{comment}
\vspace{-20pt}
\section{Experiments}
\label{sec:exp}
\begin{figure}
\floatbox[{\capbeside\thisfloatsetup{capbesideposition={left,bottom},capbesidewidth=0.38\textwidth}}]{figure}[\FBwidth]
{\caption{{\bf Bad ground truth.} Some BMS-26 videos contain significant ground truth errors, such as this segmentation of the foreground wall, which is clearly not a moving object.}\label{fig:bad_ground_truth}}
{\includegraphics[width=0.28\textwidth]{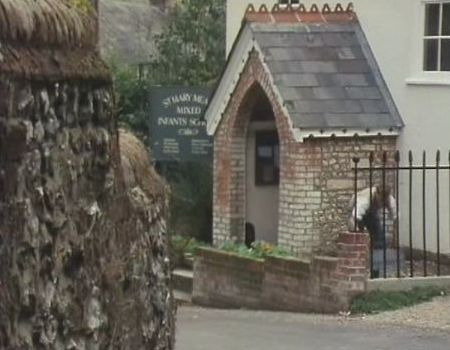}
\includegraphics[width=0.28\textwidth]{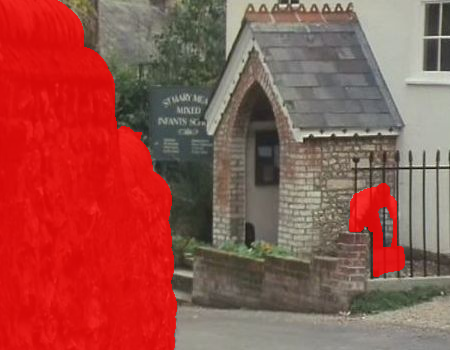}}
\end{figure}

Several motion segmentation benchmarks exist, but
  often a clear definition of what people intend to segment in 
ground truth is missing. The resulting inconsistent
  segmentations complicate the comparison of methods. We define motion
segmentation as follows. 
\vspace{10pt}
\begin{compactenum}[(I)]
%\begin{itemize}
\item Every pixel is given one of {\bf two labels}: static background or moving objects.
\item If only part of an object is moving (like a moving person with a
  stationary foot), the {\bf entire object} should be segmented.
\item {\bf All freely moving objects} (not just one) should
be segmented, but nothing else. We do not considered tethered objects such 
as trees to be freely moving.
\item Stationary objects are not segmented, even when
  they moved before or will move in the future. We consider 
  segmentation of previously moving objects to be {\em
    tracking}. Our focus is on segmentation by motion analysis.
%\end{itemize}
\end{compactenum}
\vspace{10pt}
Experiments were run on two previous datasets and our new camouflaged
animals videos.  The first was the Berkeley Motion Segmentation
(BMS-26) database ~\cite{Brox10,Tron07} (Figure~\ref{fig:transFlow}, rows
5,6). Some BMS videos have an inconsistent definition of ground truth
from both our definition and from the other videos in the benchmark.
An example is {\em Marple10} whose ground truth segments a wall in the
foreground as a moving object (see Figure~\ref{fig:bad_ground_truth}). While it is interesting to use camera motion to segment static objects (as
in~\cite{wang1994representing}), we are addressing the segmentation of
objects that are moving differently than the background, and so we
excluded ten such videos from our experiments (see Supp. Mat.).  The
second database used is the Complex Background Data Set~\cite{Manju},
which includes significant depth variation in the background and also
signficant amounts of camera rotation (Figure~\ref{fig:transFlow}, rows
3,4).  We also introduce the Camouflaged Animals Data Set
(Figure~\ref{fig:transFlow}, rows 1,2) which will be released at
camera-ready time. These videos were ground-truthed every 5th
frame. See Supp. Mat. for more.
\vspace{-10pt}
\vspace{-10pt}

{\bf Setting von Mises parameters.} 
There are two parameters $a$ and $b$ that affect the von Mises
concentration $\kappa=ar^b$. To set these parameters for each video,
we train on the remaining videos in a leave-one-out paradigm,
maximizing over the values ${0.5, 1.0, 2.0, 4.0}$ for multiplier
parameter $a$ and the values ${0, 0.5, 1, 2}$ for the exponent
parameter $b$. Cross validation resulted in the selection of the
parameter pair $(a=4.0,b=1.0)$ for most videos, and we adopted these
as our final values.

\newlength{\breite}
\breite10mm

\begin{table}
\begin{center}

\footnotesize
 \centering
\begin{tabular}
{p{0.15\textwidth}p{0.1\textwidth}z{0.11\textwidth}z{0.11\textwidth}z{0.11\textwidth}z{0.11\textwidth}z{0.11\textwidth}z{0.11\textwidth}} %{l*{8}{c}}
\toprule 
&  & {\bf Keuper} & {\bf Papaz.} & {\bf Frag.}  & {\bf Zama.} & {\bf Naray.} &{\bf ours}\\
& & {\bf \cite{keuper2015motion}} & {\bf \cite{Ferrari}}  &  {\bf \cite{fragkiadaki2012video}} & {\bf \cite{Zamalieva}} & {\bf \cite{Manju}} & \\
\midrule 

{ Camouflage} & { MCC}  & \textcolor{RoyalBlue}{\bf 0.4305} & 0.3517 &  0.1633 &  0.3354 & - & \textcolor{LimeGreen}{\textbf{0.5344}}\\                
%\cline{2-5} 
& { F}  & \textcolor{RoyalBlue}{\bf 0.4379} & 0.3297 &  0.1602 &  0.3007 & - & \textcolor{LimeGreen}{\textbf{0.5276}}\\ \addlinespace
%\hline  
  
{ BMS-26} &  { MCC}  & 0.6851 & 0.6112 &  \textcolor{RoyalBlue}{\textbf{0.7187}} &  0.6399 & - & \textcolor{LimeGreen}{\bf 0.7576} \\
%\cline{2-5}  
& { F} & \textcolor{RoyalBlue}{\bf 0.7306} & 0.6412 &  0.7276  &  0.6595 & 0.6246 & \textcolor{LimeGreen}{\textbf{0.7823}}\\ \addlinespace
%\hline

{ Complex} & {  MCC} & 0.4752  & \textcolor{RoyalBlue}{\bf 0.6359} &   0.3257 & 0.3661 & - & \textcolor{LimeGreen}{\bf 0.7491}\\
%\cline{2-5}
 &  { F}  & 0.4559 & \textcolor{RoyalBlue}{\textbf{0.6220}} &   0.3300  &  0.3297 & 0.3751 & \textcolor{LimeGreen}{\bf 0.7408}\\ \addlinespace
%\hline

{Total avg.} & MCC  & \textcolor{RoyalBlue}{\textbf{0.5737}} & 0.5375 &   0.4866 &  0.5029 & - & \textcolor{LimeGreen}{\bf 0.6918}\\
 & F  & \textcolor{RoyalBlue}{\textbf{0.5970}} & 0.5446 &   0.4911 & 0.4969 & - & \textcolor{LimeGreen}{\bf 0.6990}\\
\bottomrule
\end{tabular}
\end{center}
\caption{{\bf Comparison to state-of-the-art.} Matthew's 
correlation coefficient and F-measure for each method and data
set. The ``Total avg.'' numbers average across all valid videos.}
\label{table:numerical_results}
\end{table}

{\bf Results.} In Tab.~\ref{table:numerical_results}, we compare our
model to five different state-of-the-art
methods~\cite{Ferrari,Zamalieva,Manju,fragkiadaki2012video,keuper2015motion}.
We compared against methods for which either code was available
or that had results on either of the two public databases that we
used. However, we excluded some methods (such
as~\cite{taylor2015causal}), as their published results were less accurate
than~\cite{keuper2015motion}, to whom we compared.

Some authors have scored algorithms using the number of correctly
labeled pixels.  However, when the moving object in the foreground is
small, a method can achieve a very high score simply by marking the
entire video as background. The F-measure is also not symmetric with
respect to foreground and background, and is not well-defined when a
frame contains no foreground pixels.  Matthew's Correlation
Co-efficient (MCC) handles both of these issues, and is recommended
for scoring such binary classification problems when there is a large
imbalance between the number of pixels in each
category~\cite{powers2011evaluation}. However, in order to enable
comparison with~\cite{Manju}, and to allow easier comparison to other
methods, we also included F-measures. Table 1 shows the highest
average accuracy per data set in \textcolor{LimeGreen}{\bf green} and
the second best in \textcolor{RoyalBlue}{\bf blue}, for both the
F-measure and MCC. We were not able to obtain code for Narayana et
al.~\cite{Manju}, but reproduced F-measures directly from their paper.
The method of~\cite{fragkiadaki2012video} failed on several videos
(only in the BMS data set), possibly due to the length of these
videos. In these cases, we assigned scores for those videos by
assigning all pixels to background.

Our method outperforms all other methods by a large margin, on all
three data sets, using both measures of comparison.

\section{Analysis and Conclusions}
Conditioning our angle likelihood on the flow
magnitude is an important factor in our
method. Table~\ref{table:result} shows the detrimental effect of using
a constant von Mises concentration $\kappa$ instead of one that
depends upon flow magnitude. In this experiment, we set the
parameter $b$ which governs the dependence of $\kappa$ on $\vec{t}_r$
to 0, and set the value of $\kappa$ to maximize performance. Even with the optimum constant $\kappa$, the drop
in performance was 7\%, 5\%, and a whopping 22\% across the three data
sets.
\begin{comment}
\vspace{-10pt}
\begin{figure}
    \includegraphics[width=0.44\textwidth]{figs/table_RANSACvsnoRANSAC/RANSACvsnoRANSAC.pdf}
   \caption{\textbf{RANSAC versus no RANSAC.}  Top row: robust
     initialisation with RANSAC.  Bottom row: using Bruss and Horn's
     method directly on the entire image. Left to right: flow angles
     of translational flow, flow angles of estimated background
     translation and segmentation.} % Note that without RANSAC the
    % estimated background translation is the best fit for the car
     %instead of background.}
\label{fig:RANSAC_comp}
\end{figure}
\vspace{-10pt}
\end{comment}

\newcommand{\head}[1]{\textnormal{\textbf{#1}}}

\definecolor{myGreen}{RGB}{120,167,4}

\begin{table}[H]
\footnotesize
 \centering
\begin{tabular}[h]
{p{0.18\textwidth}z{22mm}z{24mm}z{24mm}} %{l*{8}{c}}
\toprule
%& \multicolumn{3}{c}{\head{our Method}} \\
%\cmidrule(r){2-4}
& \head{final} & \head{constant $\kappa$} & \head{no RANSAC} \\
\midrule

BMS-26 &  \textcolor{myGreen}{\textbf{0.7576}} & 0.6843 & 0.6450 \\
complex &  \textcolor{myGreen}{\textbf{0.7491}} & 0.7000 & 0.5757 \\
camouflage &  \textcolor{myGreen}{\textbf{0.5344}} & 0.3128 & 0.5176 \\
				   
\bottomrule 

\end{tabular}
\caption{ \bf Effect of RANSAC and variable $\kappa$.}
\label{table:result}
\end{table}

\begin{figure}
\floatbox[{\capbeside\thisfloatsetup{capbesideposition={right,bottom},capbesidewidth=0.5\textwidth}}]{figure}[\FBwidth]
{\caption{\textbf{RANSAC vs no RANSAC.}  Top row: robust initialisation with RANSAC.  Bottom row: using Bruss and Horn's method directly on the entire image. Left to right: flow angles of translational flow, flow angles of estimated background translation and segmentation. Note that without RANSAC the estimated background translation is the best fit for the car instead of background}\label{fig:RANSAC_comp}}
{\includegraphics[width=0.45\textwidth]{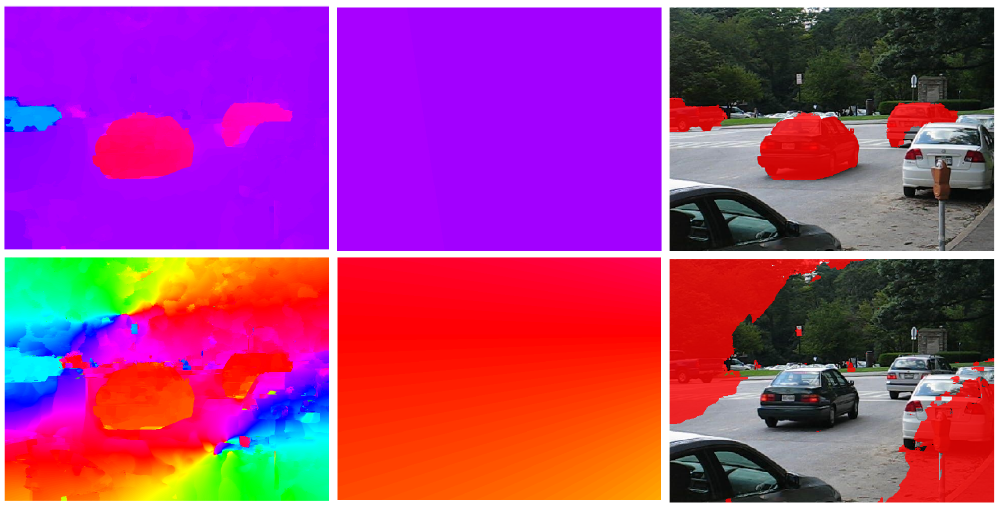}
}
\end{figure}
We also show the consistent gains stemming from
our constrained RANSAC initialization procedure. In this experiment, 
we segmented the first frame of video without rejecting any pixels as
outliers. In some videos, this had little effect, but sometimes the effect
was large, as shown in Figure~\ref{fig:RANSAC_comp}.

\begin{figure}
   \includegraphics[width=1\columnwidth]{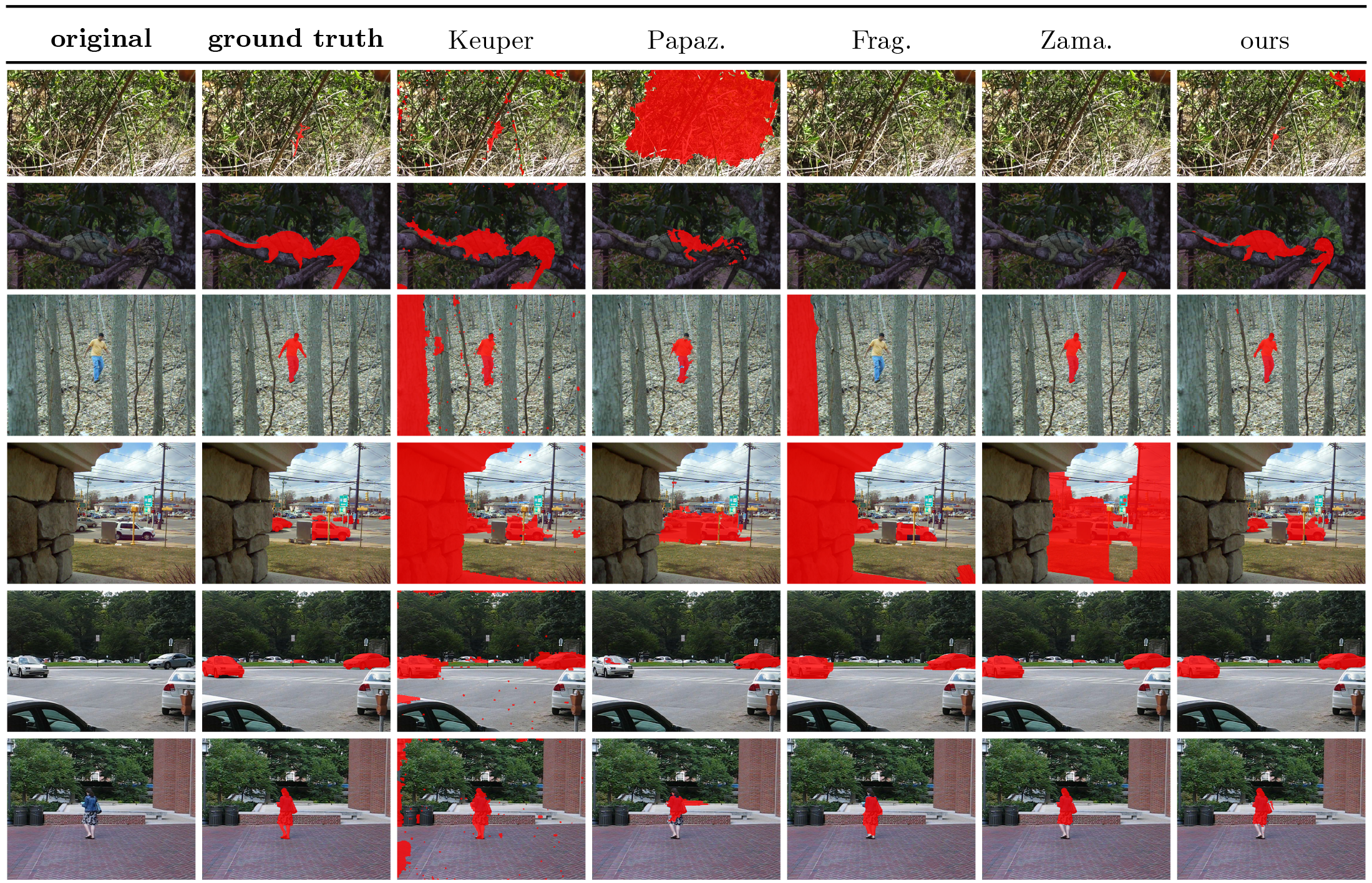}
	\caption{{\bf Sample results} Left to right: original image, ground truth, \cite{keuper2015motion}, \cite{Ferrari}, \cite{fragkiadaki2012video} \cite{Zamalieva} and our binary segmentations. Rows 1-2: sample results on the Animal Camouflage Data Set (chameleon and stickinsect). Rows 3-4: sample results on Complex Background (traffic and forest). Rows 5-6: sample results on BMS-26 (cars5 and people1).}
	\label{fig:transFlow}
\end{figure}

The method by Keuper et al.~\cite{keuper2015motion} performs fairly
well, but often makes errors in segmenting rigid parts of the
foreground near the observer. This can be seen in the third and fourth
rows of Figure~\ref{fig:transFlow}, which shows sample results from the
Complex Background Data Set. In particular, note that Keuper et al.'s
method segments the tree in the near foreground in the third row and
the wall in the near foreground in the fourth row.  The method of
Fragkiadaki et al., also based on trajectories, has similar behavior. 
These methods in general seem to have difficulty with high variability
in depth.

Another reason for our excellent performance may be that we are not making
any compromises in modeling motion. We are directly using the
perspective projection equations to analyze motion, as has been
advocated by Horn~\cite{horn1999projective}, rather than
approximations based on projective geometry, as done by Zamalieva et
al. Code is available: \url{http://vis-www.cs.umass.edu/motionSegmentation/}.

\clearpage

\bibliographystyle{splncs}
\bibliography{eccv2016submission}
\newpage
\section*{\centering Supplementary Material}
\vspace{10pt}
Our supplementary material contains:
\begin{itemize}
\item a review of the BMS-26 data set and a detailed overview about excluded video sequences, %results and evaluation on the entire datasets including videos with groundtruth which are not directly related to our understanding of motion segmentation,
\item additional details about the \textit{Camouflaged Animals Data Set},
%\item six example videos of the three data sets showing our results and the results of four \cite{Zamalieva, Ferrari, fragkiadaki2012video, keuper2015motion} competing methods,
\item related work about camera motion estimation and
\item a description of our modified Bruss and Horn error~\cite{bruss1983passive}, a fundamental part of our motion segmentation initialization.
\end{itemize}

\subsection*{Comparability of the motion segmentation data sets}
A lot of different databases have been created to provide a common benchmark for motion segmentation. Often a clear definition of what people intend to segment when they provide a ground truth is missing. This results in many inconsistent segmentations, which makes it hard to compare against other motion segmentation methods.
In our paper we give a clear definition of motion segmentation
\newline
\begin{compactenum}[(I)] %[label=\Roman*)]
\item Every pixel is given one of \textbf{two labels}: static background or moving objects.
\item If only part of an object is moving, the \textbf{entire object} should be segmented.
\item \textbf{All freely moving objects} should be segmented, but nothing else.
\item Stationary objects are not segmented, even when they moved before or will move in future. We consider segmentation of previously moving objects to be tracking.
\end{compactenum}
\vspace{1em}

In the following subsection we give detailed information about all videos that do not correspond to our understanding of motion segmentation.

\subsection*{Berkeley Motion Segmentation database (BMS-26)}

To satisfy the first criterion of motion segmentation we converted the given ground truth into a binary segmentation, removing the provided motion labels. If all four criteria are satisfied, we used the video for comparison. The effect of the mislabeled ground truth varies a lot. The difference between a correct ground truth of \textit{marple2} and the provided ground truth for example is enormous, whereas the difference of a correct ground truth of \textit{tennis} would be almost not noticeable. Trying to be as objective as possible we excluded all videos where one of our four criteria of motion definition is violated, indepently of the size of the mislabeled region. The following table shows the sequences we excluded for evaluation.

\begin{longtable}
 {m{2cm}m{3.8cm}m{0.8cm}m{0.4cm}m{0.4cm}m{0.4cm}c} 
 video & ground truth & \rom{1} & \rom{2} &  \rom{3} & \rom{4} & comment \\
\hline \hline
cars9 & \includegraphics[width=0.12\textwidth]{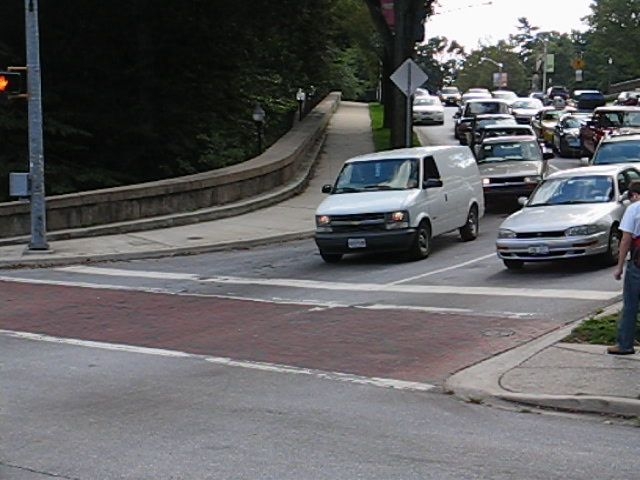}
\includegraphics[width=0.12\textwidth]{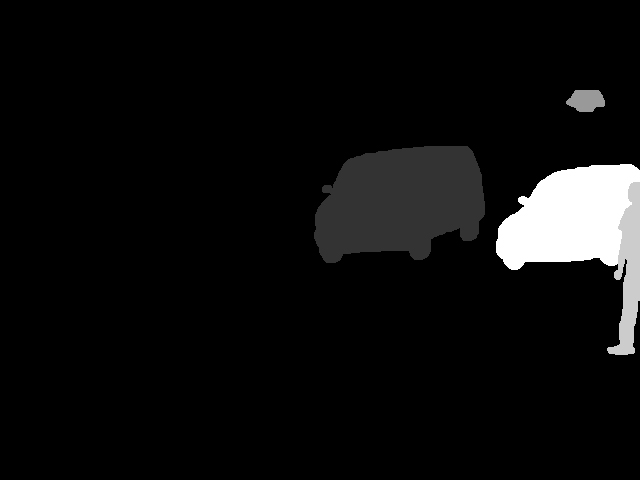}  & 5 & \cmark & \cmark &  \xmark &  \multirow {1}{3.2cm}{\footnotesize white van is segmented before it starts to move} \\
people2 & \includegraphics[width=0.12\textwidth]{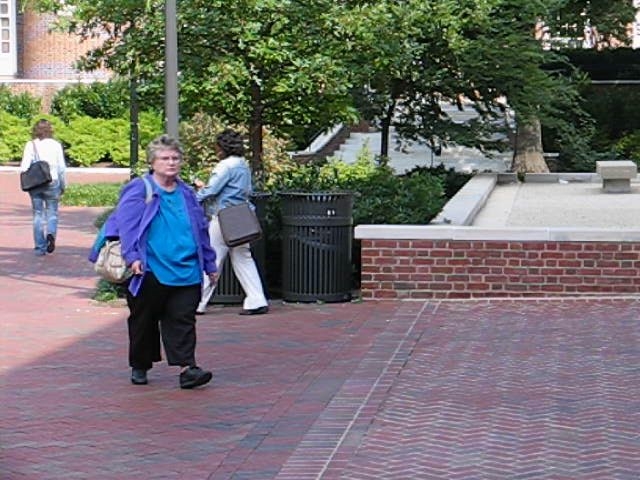}
\includegraphics[width=0.12\textwidth]{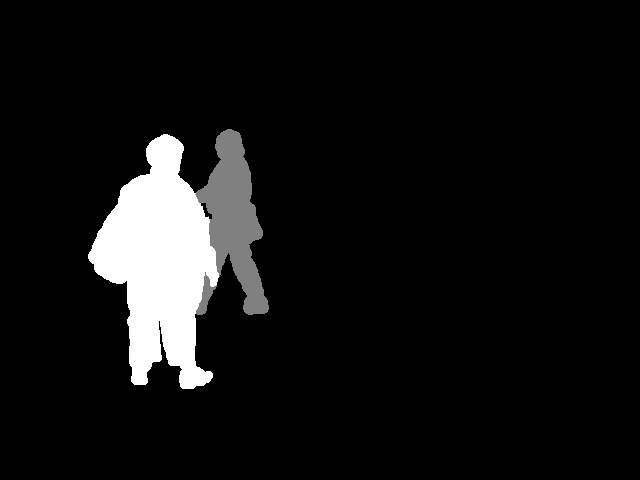}  & 3 & \cmark & \xmark &  \cmark & \multirow {1}{3.3cm}{\footnotesize third person in the bg is not segmented}  \\
tennis & \includegraphics[width=0.12\textwidth]{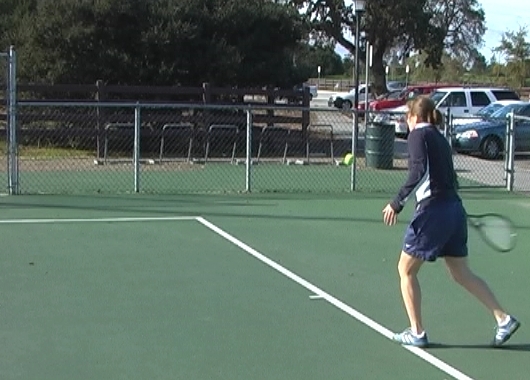}
\includegraphics[width=0.12\textwidth]{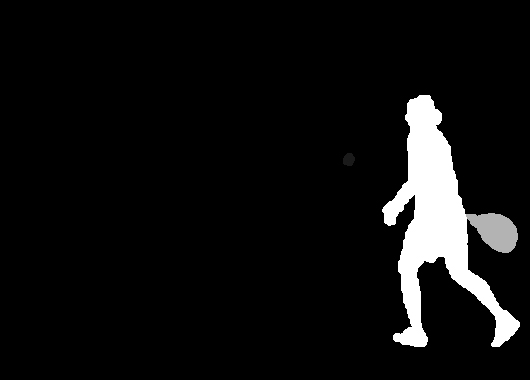}  & 4 & \cmark & \xmark &  \cmark & \multirow {1}{3.2cm}{\footnotesize tennis ball is not segmented}  \\
marple2 & \includegraphics[width=0.12\textwidth]{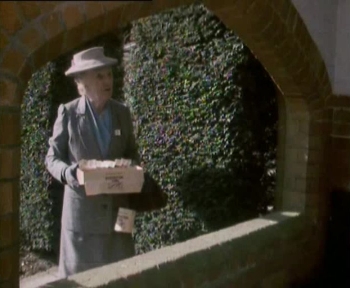}
\includegraphics[width=0.12\textwidth]{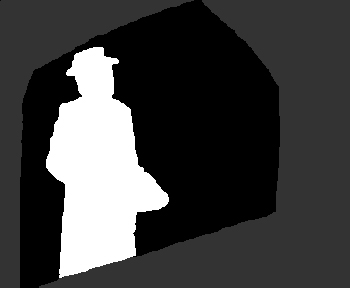} & 5 & \cmark & \xmark &  \cmark & \multirow {1}{3.2cm}{\small static fg building is segmented}  \\
marple6 & \includegraphics[width=0.12\textwidth]{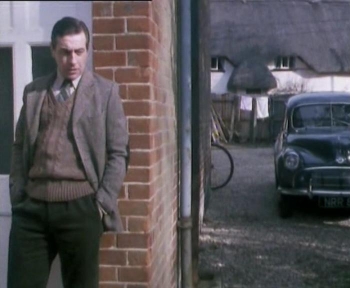}
\includegraphics[width=0.12\textwidth]{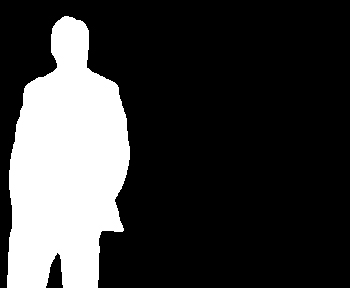}  & 3 & \cmark & \xmark &  \xmark & \multirow {1}{3.2cm}{\footnotesize bike in the bg is not segmented}  \\
marple7 & \includegraphics[width=0.12\textwidth]{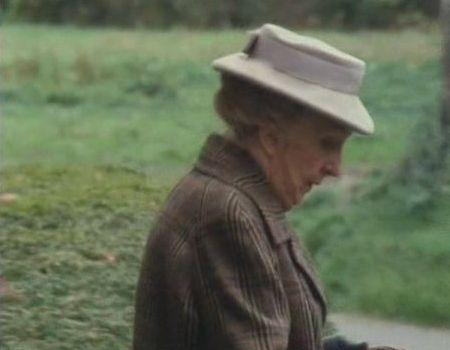}
\includegraphics[width=0.12\textwidth]{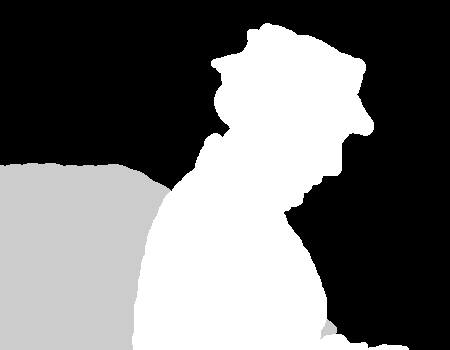} & 5 & \cmark & \xmark &  \xmark & \multirow {1}{3.2cm}{\footnotesize static hedge is segmented}  \\
marple10 & \includegraphics[width=0.12\textwidth]{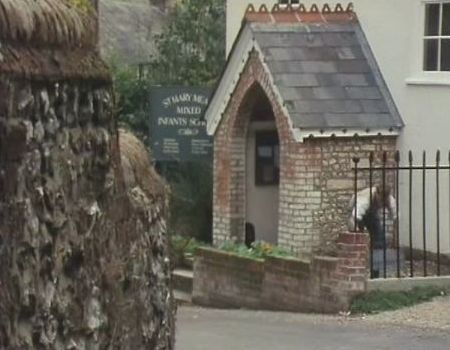}
\includegraphics[width=0.12\textwidth]{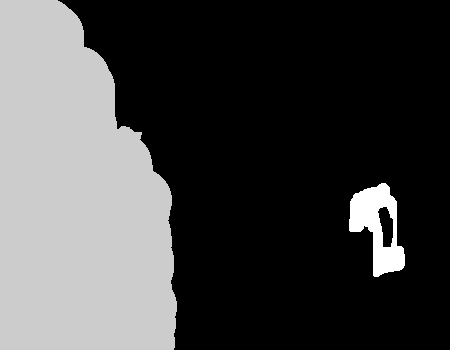} & 5 & \cmark & \xmark &  \cmark & \multirow {1}{3.2cm}{\footnotesize static wall is segmented}  \\
marple11 & \includegraphics[width=0.12\textwidth]{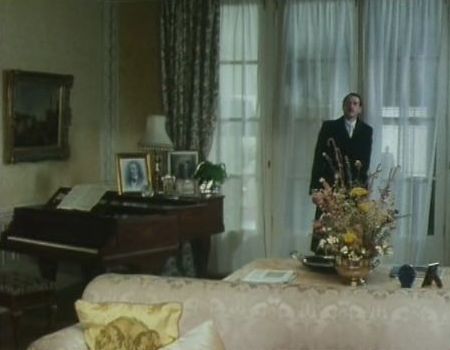}
\includegraphics[width=0.12\textwidth]{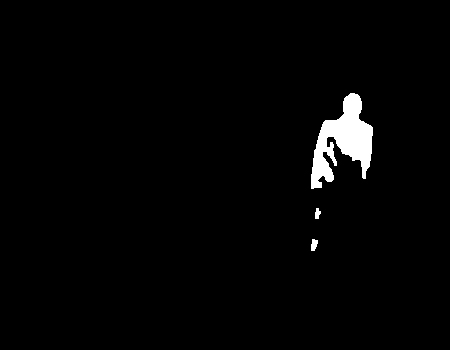} & 2 & \cmark & \cmark &  \xmark & \multirow {1}{3.2cm}{\footnotesize man is segmented before he starts to move}  \\
marple12 & \includegraphics[width=0.12\textwidth]{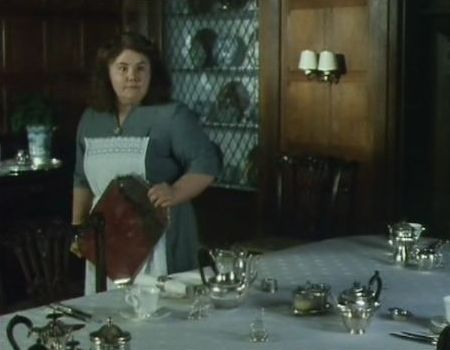}
\includegraphics[width=0.12\textwidth]{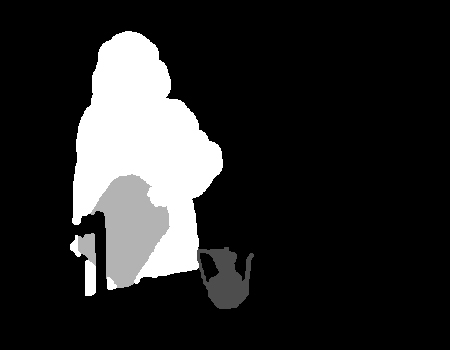}  & 4 & \cmark & \cmark &  \xmark & \multirow {1}{3.2cm}{\footnotesize pot is segmented, which was moved before}  \\
marple13 & \includegraphics[width=0.12\textwidth]{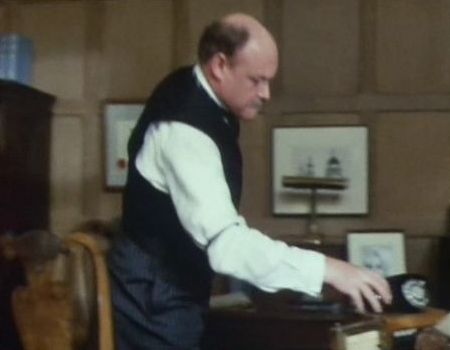}
\includegraphics[width=0.12\textwidth]{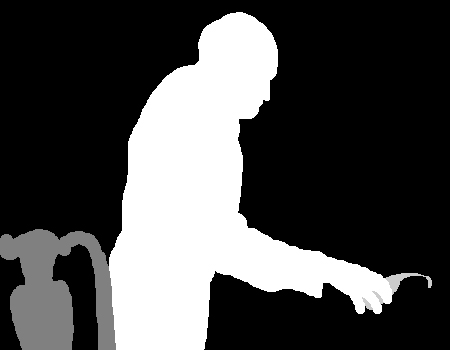} & 4 & \cmark & \cmark &  \xmark & \multirow {1}{3.2cm}{\footnotesize chair is segmented, which was moved before}\\
\caption{{\bf Ground truth of BMS-26} video sequences we excluded for evaluation due to mislabeled regions in ground truth.}
\label{table:comparison}
\end{longtable}

\subsection*{Camouflaged Animals Data Set}
Our new data set includes nine short video sequences extracted from
YouTube videos and an accompanying ground truth. Table 1 shows the link
to the original YouTube video, the exact time where our chosen video
sequence starts within the YouTube video, and the number of frames the
sequence contains. For our motion segmentation algorithm, we converted
the video sequence to an image sequence in the png format using the
VideoReader function of Matlab. Each sequence contains hand-labeled
ground truth of moving objects in every fifth frame.

\begin{center}
    \begin{tabular}{ | c | c | c | c|}
    \hline
    Video  & Link  & Start & Frames \\ \hline
chameleon & \cite{utube_camouflage_1} & 02:28.20 & 218 \\
frog & \cite{utube_camouflage_2} & 00:05.38 & 31 \\
glow-worm beetle & \cite{utube_camouflage_3} & 06:02.13 & 104 \\
snail & \cite{utube_camouflage_3} & 06:07.02 & 84\\
scorpion1 & \cite{utube_camouflage_3} & 02:09.00 & 105 \\
scorpion2 & \cite{utube_camouflage_3} & 02:25.04 & 61 \\
scorpion3 & \cite{utube_camouflage_3} & 02:27.48 & 76\\
scorpion4 & \cite{utube_camouflage_3} & 00:06.11 & 80 \\
stickinsect & \cite{utube_phasmid} & 00:05.68 & 80 \\
\hline
\end{tabular}
\end{center}

\subsection*{Camera motion estimation in the context of motion segmentation}

In camera motion estimation, also known as egomotion estimation, the
relationship between the observed optical flow and the camera's
translation and rotation are exploited in different ways to estimate
the six motion parameters.  Jepson and Heeger~\cite{Jepson94}
construct a set of constraints for optical flow observations in a
manner that effectively negates the rotation component.
Prazdny~\cite{Prazdny83} shows that the difference between any two
flow vectors yields a constraint that does not depend on rotation.
Tomasi and Shi~\cite{Tomasi93} estimate translation by using image
deformation rather than movement of points within an image, to
minimize the effects of rotation. In the above methods, translation is
first estimated using the rotation invariant constraints, followed by
rotation estimation.  Prazdny~\cite{Prazdny80} estimates the rotation
first by using constraints that are independent of camera translation
and the depth of the points in the scene. These and other methods for
egomotion estimation are described by Tian et al.~\cite{Tian96}.
Recently, many authors (e.g.~\cite{Yamaguchi13}) have used projective
geometry methods such as the 8-point algorithm~\cite{hartley1997defense} in combination with RANSAC~\cite{RANSAC} for robustness.

We are not directly interested in computing the egomotion, but rather
in transforming the observed optical flow vectors such that the flow
vectors are a function of only translation, and not rotation.  This is
similar in spirit to Yamaguchi et al.~\cite{Yamaguchi13}, who estimate
the rotation at each pixel such that when the rotational flow vectors
are subtracted from the observed flow vectors, a pure translational
flow field is obtained.  But they do this under the assumption of 
a static scene. We demonstrate successful motion modeling in the presence
of multiple large moving objects in Section 3 of our paper. Based on a least squares minimization to find the camera translation ~\cite{bruss1983passive} we slightly modify the error function to apply the camera motion estimation problem to the motion segmentation problem. Furthermore we add a RANSAC procedure as initialization for robustness and outlier reduction.

\subsection*{Modified Bruss and Horn Error}
As described in the main text, we introduced a modification to the 
error function of the Bruss and Horn algorithm that we call the modified
Bruss and Horn (MBH) error. We first give some basic background on 
perspective projection, describe the Bruss and Horn error function and
a particular issue that makes it problematic in the context of motion
segmentation, and then describe our modification to the algorithm.

{\bf Basics of perspective projection.} Given a particular set of translation parameters $(U,V,W)$ (and assuming
no camera rotation), the direction of optical flow of the background can
be predicted for each point $(x,y)$ in the image
via the perspective projection equations.
Let a point $P$ have 3-D coordinates $(X,Y,Z)$. Then the image coordinates $(x,y)$
for this point are:
\begin{equation}
x=\frac{Xf}{Z}
\qquad
\text{and}
\qquad
y=\frac{Yf}{Z},
\end{equation}
where $f$ is the camera's focal length. 
A tranlational camera motion $(U,V,W)$ yields in a pixel displacment $\vec{v}$ in the image.
\begin{equation}
v_x=\frac{W \cdot x - U \cdot f}{Z}
\qquad
\text{and}
\qquad
v_y=\frac{W \cdot y - V \cdot f}{Z}.
\end{equation}
The {\em direction} of the motion, given by 
\begin{equation}
\arctan(W \cdot y - V \cdot f, W \cdot x - U \cdot f),
\end{equation} 
is then a function of the original image position $(x,y)$, the
direction of motion $(U,V,W)$ and the focallength $f$, and has no dependence on the depth $Z$ of the
point.

\vspace{-10pt}
\begin{figure}
\floatbox[{\capbeside\thisfloatsetup{capbesideposition={left,top},capbesidewidth=6cm}}]{figure}[\FBwidth]
{\caption{{\bf Bruss and Horn error.} Let $\vec{p}$ be a vector in the direction of preferred motion with respect to a motion hypothesis $(U,V,W)$. The Bruss and Horn error assigned to a translational flow vector $\vec{v}_t$ is then the distance of its projection onto $\vec{p}$. However, this {\em same error} would be assigned to a vector $-\vec{v}_t$ pointing in the opposite direction, which should have much lower compatibility with the motion hypothesis.}\label{fig:BrussHornError}}
{\includegraphics[width=0.95\linewidth]{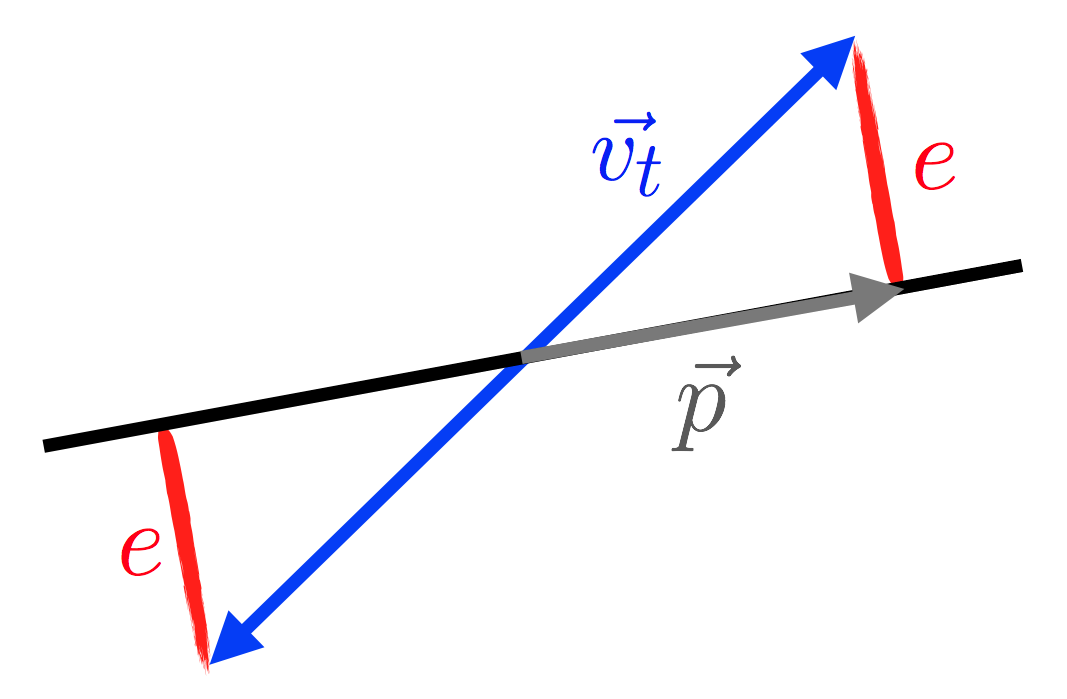}}
\end{figure}
\vspace{-10pt}
{\bf The Bruss and Horn Error Function.} The point of the Bruss and
Horn algorithm (translation-only case) is to find the motion direction
parameters $(U,V,W)$ that are as compatible as possible with the
observed optical flow vectors.  Let $\vec{p}$ be a vector in the
direction of the flow expected from a motion $(U,V,W)$ 
(see Figure~\ref{fig:BrussHornError}). Then the Bruss
and Horn error for the observed flow vector $\vec{v}_t$ is the
distance of the projection of $\vec{v}_t$ onto $\vec{p}$, shown by
the red segment $e$ on the right side of the figure. 

The problem with this error function is that this distance is small
not only for vectors which are close to the preferred direction, but
also for vectors that are in a direction {\em opposite} the preferred
direction. That is, {\em observed optical flow vectors that point in
  exactly the wrong direction with respect to a motion $(U,V,W)$ get a
  small error} in the Bruss and Horn algorithm. In particular, the error
assigned to a vector $\vec{v}_t$ is the same as the error assigned to a
vector $-\vec{v}_t$ in the opposite direction (See Figure~\ref{fig:BrussHornError}).

Because the Bruss and Horn algorithm is intended for motion estimation
in scenarios where there is only a single moving object (the
background), such motions in the opposite direction to the preferred
motion are not common, and thus, this ``problem'' we've identified has
little impact. However, in the motion segmentation setting, where
flows of objects may be in opposite directions, this can make the flow
of a separately moving object (like a car), look as though it is
compatible with the background.  We address this problem by
introducing a modified version of the error.

{\bf The modified Bruss and Horn error.} 
As stated above, the Bruss and Horn error is the distance of the projection
of an optical flow vector onto the vector $\vec{p}$ representing the
preferred direction of flow according to a motion model $(U,V,W)$. This
can be written simply as 
\begin{equation}
e_{BH}(\vec{v}_t,\vec{p})=\Vert \vec{v}_t \Vert \cdot \lvert \sin(\measuredangle(\vec{v}_t,\vec{p})\rvert .
\end{equation}

This error function has the appropriate behavior when the observed optical
flow is within 90 degrees of the expected flow direction, i.e., when
$\vec{v}_t \cdot \vec{p} \geq 0.$ However, when the observed flow points
{\em away} from the preferred direction, we assign an error equal to the
magnitude of the entire vector, rather than its projection, since no
component of this vector represents a ``valid direction'' with respect
to $(U,V,W)$. This results in the modified Bruss and Horn error (see Figure~\ref{fig:ModifiedBrussHornError}):
\vspace{-10pt}
\begin{figure}
\floatbox[{\capbeside\thisfloatsetup{capbesideposition={left,top},capbesidewidth=6cm}}]{figure}[\FBwidth]
{\caption{{\bf Modified Bruss and Horn error.} When an observed
     translation vector $\vec{v}_t$ is within 90 degrees of the
     preferred direction, its error is computed in the same manner as the
     traditional Bruss and Horn error (right side of figure). However,
     when the observed vector is more than 90 degrees from the preferred
direction, its error is computed as its full magnitude, rather than the distance of projection (left side of figure). This new error function keeps objects moving in opposite directions from being confused with each other.
}\label{fig:ModifiedBrussHornError}}
{\includegraphics[width=0.95\linewidth]{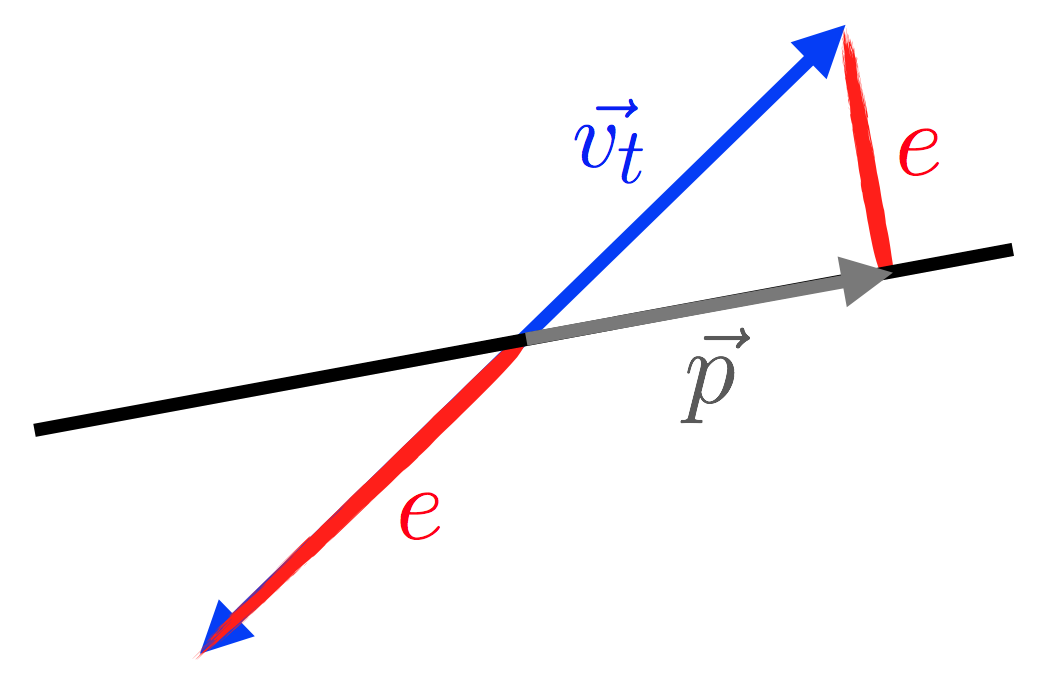}}
\end{figure}
\vspace{-10pt}
\begin{equation}
    e_{MBH}= 
\begin{cases}
    \Vert \vec{v}_t \Vert ,& \text{if } \vec{v}_t \cdot \vec{p} < 0\\
    \Vert \vec{v}_t \Vert \cdot \lvert \sin(\measuredangle(\vec{v}_t,\vec{p})\rvert,              & \text{otherwise.}
\end{cases}
\end{equation}
This error has the desired behavior of penalizing flows in the opposite
direction to the expected flow.

\end{document}